\documentclass[sigconf]{acmart}
\renewcommand\footnotetextcopyrightpermission[1]{} 
\usepackage{threeparttable}
\usepackage{subcaption}
\acmConference[]{}

\renewcommand\footnotetextcopyrightpermission[1]{}
\setcopyright{none}

\AtBeginDocument{%
  }

\usepackage{siunitx}
\begin{document}

\title{Quantization of Spiking Neural Networks Beyond Accuracy}

\author{Evan Gibson Smith}
\email{egsmith@wpi.edu}
\affiliation{%
  \institution{Worcester Polytechnic Institute}
  \city{Worcester}
  \state{Massachusetts}
  \country{USA}
}




\author{Jacob Whitehill}
\email{jrwhitehill@wpi.edu}
\affiliation{%
  \institution{Worcester Polytechnic Institute}
  \city{Worcester}
  \state{Massachusetts}
  \country{USA}
}

\author{Fatemeh Ganji}
\email{fganji@wpi.edu}
\affiliation{%
  \institution{Worcester Polytechnic Institute}
  \city{Worcester}
  \state{Massachusetts}
  \country{USA}
}





\renewcommand{\shortauthors}{Smith et al.}
\acmArticleType{Research}
\acmContributions{EGS designed and conducted the experiments, analyzed the results, and wrote the manuscript. JW and FG supervised the research and provided guidance throughout.}



\begin{abstract}
Quantization is a natural complement to the sparse, event-driven computation 
of Spiking Neural Networks, reducing memory bandwidth and arithmetic cost for 
deployment on resource-constrained hardware. However, existing SNN 
quantization evaluation focuses almost exclusively on accuracy, overlooking 
whether a quantized network preserves the firing behavior of its 
full-precision counterpart. We demonstrate that quantization method, clipping 
range, and bit-width can produce substantially different firing distributions 
at equivalent accuracy, differences invisible to standard metrics but 
relevant to deployment, where firing activity governs effective sparsity, 
state storage, and event-processing load. To capture this gap, we propose 
Earth Mover's Distance as a diagnostic metric for firing distribution 
divergence, and apply it systematically across weight and membrane 
quantization on SEW-ResNet architectures trained on CIFAR-10 and CIFAR-100. 
We find that uniform quantization induces distributional drift 
even when accuracy is preserved, while LQ-Net style learned quantization maintains firing 
behavior close to the full-precision baseline. Our results suggest that 
behavior preservation should be treated as an evaluation criterion 
alongside accuracy, and that EMD provides a principled tool for assessing it.
\end{abstract}

\maketitle

\section{Introduction}

Spiking Neural Networks (SNNs) have emerged as a compelling paradigm 
for energy-efficient inference, leveraging sparse, event-driven computation 
to reduce power consumption compared to conventional artificial neural 
networks~\cite{davies2018loihi}. Modern architectures such as the Spiking 
Element-Wise ResNet (SEW-ResNet)~\cite{fang2021deep} have narrowed the 
accuracy gap with ANNs while retaining the fundamental efficiency advantages 
of spike-based processing. As SNNs transition from research prototypes to 
deployed systems, model compression techniques become essential for edge and 
neuromorphic deployment.

Quantization is a natural complement to SNNs' inherent sparsity. Lower-bit 
representations decrease memory bandwidth, arithmetic cost, and storage 
requirements, enabling deployment on resource-constrained hardware. However, 
quantization in SNNs introduces challenges absent in conventional networks. 
Unlike ANNs where activations are continuous-valued, SNNs produce binary spike 
trains whose statistics (firing rates, temporal patterns, and population 
dynamics) directly govern network behavior. A quantized SNN that preserves task 
accuracy may nonetheless exhibit substantially altered firing dynamics: shifts 
in sparsity structure, changes in dead neuron populations, and distributional 
drift from the full-precision baseline. These differences are invisible to 
accuracy metrics but relevant to deployment, where firing activity influences 
effective sparsity, state storage, and event-processing load.

Existing SNN quantization work focuses primarily on accuracy preservation, 
leaving the question of \textit{behavior preservation} underexplored. We argue 
that accuracy alone is an insufficient criterion: two quantized models with 
similar accuracy may differ substantially in their firing distributions, with 
direct consequences for how they behave in deployment. To capture this gap, we 
propose using the Earth Mover's Distance (EMD) as a diagnostic metric, 
measuring distributional divergence between a quantized SNN and its 
full-precision baseline. EMD is particularly suited to this setting because 
SNN firing distributions are often non-normal, exhibiting heavy tails or 
multi-modal structure that scalar summary statistics fail to capture.

We apply this framework systematically across quantization methods, bit-widths, 
clipping ranges, and architectures, evaluating both weight and membrane 
quantization on CIFAR-10 and CIFAR-100. Our contributions are as follows:

\begin{itemize}
    \item \textbf{EMD as an evaluation criterion for SNN quantization:} We 
    demonstrate that accuracy alone is insufficient to evaluate quantized SNNs, 
    and introduce Earth Mover's Distance as a principled diagnostic metric for 
    measuring firing distribution divergence from full-precision baselines.
        
    \item \textbf{Systematic characterization of quantization effects on spiking 
    dynamics:} We show that quantization method, clipping range, and bit-width 
    produce substantially different firing distributions at equivalent accuracy. 
    We apply LQ-Net-style learned quantization to SNNs for the first time, and 
    find that learned quantization preserves firing rate behavior where uniform quantization 
    fails. Membrane quantization presents distinct challenges requiring higher 
    precision and unsigned ranges.

\end{itemize}

\section{Related Work}

Spiking neural networks can be trained either through ANN-to-SNN 
conversion or through direct training with surrogate gradients. 
Conversion methods such as QFFS~\cite{li2022qffs}, which achieves 
70.18\% on ImageNet in 8 timesteps, quantize ANN activations prior 
to conversion but operate in a fundamentally different setting from 
direct training. We focus on direct training, which has been shown 
to be more performant under fewer timesteps and more constrained 
network configurations. SNNs are a prime candidate for 
resource-constrained deployment due to their sparse, event-driven 
computation on specialized hardware such as Loihi~\cite{davies2018loihi} 
and TrueNorth~\cite{akopyan2015truenorth}.

Prior work on SNN quantization focuses primarily on accuracy
recovery. Q-SNNs~\cite{wei2024qsnns} quantize both weights and
membrane potentials using a regularization strategy that maximizes
information entropy of weight and spike distributions, recovering
accuracy at low bit-widths. While Q-SNNs do visualize spike
distributions to validate their regularization, they do so only to
confirm accuracy recovery. The distributional pathologies induced
by quantization are not characterized, and no metric is proposed
for measuring dynamics divergence from a full-precision baseline.
QP-SNN~\cite{wei2024qpsnn} combines uniform weight quantization with
structured pruning, noting that naive uniform quantization leads to
inefficient bit usage. SpikeFit~\cite{SpikeFit_Kartashov_Pushkareva_Karandashev_2025} applies  Clusterization-Aware Training (CAT) to SNNs, learning a small set of  discrete codebook values for weights through clustering. In contrast, LQ-Net learns a set of 
basis vectors via closed-form error minimization, with quantized weights expressed as binary linear combinations of those vectors. SpQuant-SNN~\cite{hassan2024spquant}
introduces a learnable membrane threshold scalar to improve
quantization robustness; crucially, this learns a single scalar
per layer rather than full quantization basis vectors, and no
analysis of firing dynamics is provided. In contrast to all of
these, we characterize the distributional consequences of
quantization choices and propose EMD as a principled diagnostic
for dynamics preservation, independent of accuracy.
Quantization methods broadly split into post-training quantization
(PTQ) and quantization-aware training (QAT). QAT consistently
outperforms PTQ at low bit-widths by simulating quantization during
training, typically using the Straight-Through
Estimator~\cite{bengio2013estimating} to approximate gradients
through the rounding operation. Within QAT, learned quantization
methods go beyond fixed uniform levels. LQ-Net~\cite{zhang2018lq}
jointly optimizes a set of basis vectors with network weights via a
closed-form Quantization Error Minimization step, learning
non-uniform levels adapted to the weight distribution.
LSQ~\cite{esser2020lsq} takes a simpler approach, learning a single
scale factor per layer via backpropagation. Both methods were
developed for ANNs. While SpikeFit~\cite{SpikeFit_Kartashov_Pushkareva_Karandashev_2025} applies learned codebook levels
to SNNs in a hardware-constrained setting, we apply LQ-Net-style
basis vector quantization to SNNs for the first time in the context
of dynamics preservation, evaluating its effect on firing
distribution divergence from full-precision baselines. While
learnable quantization thresholds across timesteps have been
explored in \textit{Trainable Quantization for Speedy Spiking
Neural Networks}~\cite{castagnetti2023trainable}, similar in spirit
to SpQuant-SNN, these approaches learn scalar parameters rather
than full quantization basis vectors.

For our backbone, we use SEW-ResNet~\cite{fang2021deep} with ADD 
residual connections, which enables stable training of deep spiking 
networks. In the context of spiking residual networks, SynA-ResNet~\cite{synaresnet2023} is notable in our 
context for observing that training can induce emergent dead neurons. To measure distributional divergence in firing behavior, we use the Earth Mover's Distance (Wasserstein-1 
distance)~\cite{rubner2000emd}, which captures shape changes such 
as bimodality and variance shifts that scalar summary statistics 
miss. While EMD has been applied to temporal spike train 
distances~\cite{EMD_SpikeTrain_Sihn_Kim_2019}, we apply it to per-neuron firing rate distributions, a use that is, to our knowledge, novel in the SNN quantization setting.

\section{Methods}

\subsection{Model Architecture and Training Setup}

We evaluate quantization on the Spiking Element-Wise ResNet 
(SEW-ResNet)~\cite{fang2021deep}, specifically the ResNet-8 and 
ResNet-18 variants, trained directly on CIFAR-10 and CIFAR-100. 
These architectures use ADD-type spike-element-wise residual 
connections and Leaky Integrate-and-Fire (LIF) neurons with 
$\tau = 2.0$, $v_{\text{threshold}} = 1.0$, $v_{\text{reset}} = 0.0$, 
and $T = 4$ timesteps with direct encoding. We select these 
architectures for their suitability for edge and neuromorphic 
deployment, and because their moderate scale permits systematic 
sweep across quantization methods and bit widths.

All models are trained from scratch using SGD with momentum $0.9$, 
weight decay $5 \times 10^{-4}$, initial learning rate $0.1$, and 
cosine annealing over 200 epochs with $\eta_{\min} = 10^{-4}$. 
Batch size is 128. We use standard data augmentation for CIFAR 
(random crop with padding 4, random horizontal flip).

\subsection{Quantization Methods}

We evaluate quantization across two axes: \emph{what} is quantized 
(weights vs.\ membrane potentials) and \emph{how} it is quantized 
(quantization scheme). This yields a systematic comparison across 
the following methods.

\subsubsection{Uniform Quantization with Straight-Through Estimator}

As our primary baseline, we apply symmetric uniform quantization to 
network weights using the Straight-Through Estimator 
(STE)~\cite{bengio2013estimating} for gradient approximation during 
QAT. Weights are clipped to $[\text{clip\_min}, \text{clip\_max}] = 
[-10, 10]$ and $[\text{clip\_min}, \text{clip\_max}] = 
[-1, 1]$  and uniformly discretized into $2^n$ levels. The 
quantization function is:
\begin{equation}
    \hat{w} = \text{round}\!\left(
        \frac{w - w_{\min}}{w_{\max} - w_{\min}} \cdot (2^n - 1)
    \right) \cdot \frac{w_{\max} - w_{\min}}{2^n - 1} + w_{\min}
\end{equation}
where gradients pass through the rounding operation unchanged 
during backpropagation. We evaluate at $n \in \{2, 4, 6, 8\}$ bits.

\subsubsection{Learned Quantization (LQ-Net)}

We apply LQ-Net~\cite{zhang2018lq} to SNNs, to the best of our 
knowledge for the first time. LQ-Net replaces fixed uniform levels 
with a set of $n$ learned basis vectors $\mathbf{b} \in 
\mathbb{R}^n$, expressing quantized weights as:
\begin{equation}
    \hat{w} = \sum_{k=1}^{n} b_k \cdot q_k, \quad q_k \in \{0, 1\}
\end{equation}
where $\{q_k\}$ are binary codes and $\mathbf{b}$ is jointly 
optimized with network weights via a closed-form Quantization Error 
Minimization (QEM) step during each forward pass. Basis vectors are 
initialized from the expected magnitude of normally distributed 
weights scaled to the layer fan-in, and updated using a moving 
average for stability. For weight quantization, we apply per-channel 
quantizers with symmetric initialization. We evaluate LQ-Net QAT 
at $n \in \{2, 4, 6, 8\}$ bits.

\subsection{Membrane Potential Quantization}

In addition to weight quantization, we investigate quantization of 
neuron membrane potentials, which provide another avenue for 
compression relevant to hardware memory bandwidth. Membrane potentials, as continuous-valued recurrent state variables, pose different quantization challenges from weights. The distribution of membrane potentials varies across timesteps and thus quantization noise can compound across timesteps. They are also not stationary like weights, depending on the particular input, much like activations in standard neural networks.

We apply uniform quantization to membrane potentials within the LIF 
update rule, discretizing $v[t] \in [\text{clip\_min}, 
\text{clip\_max}]$ prior to the threshold comparison at each 
timestep. Due to the variance in membrane distributions across batches and timesteps, we find that a naive LQ-Net-style learned membrane quantization is unable to train. Based on this observation, modifications to enable LQ-Net style membrane quantization we leave as an open challenge.

\subsection{Dynamics Analysis}

A central contribution of this work is characterizing the effect 
of quantization on \emph{spiking dynamics}, beyond accuracy alone. 
For each trained model, we collect per-neuron firing rates by 
running inference over 20 batches of the test set and averaging 
spike outputs across timesteps and spatial dimensions, yielding 
a scalar firing rate $r_c \in [0, 1]$ for each channel $c$.

We then compute the following metrics relative to the 
full-precision baseline:

\textbf{Mean firing rate.} The mean of $\{r_c\}$ across all 
channels in a layer, averaged across layers for a network-level 
summary.

\textbf{Earth Mover's Distance (EMD).} The Wasserstein-1 distance 
between the firing rate distribution of the quantized model and 
the full-precision baseline:
\begin{equation}
    \text{EMD} = W_1\!\left(P_{\text{quant}},\, 
    P_{\text{FP32}}\right)
\end{equation}
where $P$ denotes the empirical distribution of per-neuron firing 
rates within a layer. EMD captures not only shifts in mean firing 
rate but distributional shape changes such as bimodality, 
variance shifts, and the emergence of dead or saturated neuron 
populations. We compute EMD between two networks across each layer, and report network averages.

\textbf{Dead and saturated neuron percentage.} We report the 
fraction of neurons with $r_c < 0.05$ (dead) and mean neuron firing rate as indicators of pathological firing regimes.

\begin{table*}[h!]
\caption{Weight quantization comparison across clipping ranges and architectures.
At higher bit-widths, wide clipping ([-10,10]) achieves accuracy comparable to narrow
([-1,1]) but produces substantially higher EMD, revealing pathological firing dynamics
invisible to accuracy alone. LQ-Net preserves dynamics across all configurations.
At 2-bit with wide clipping, uniform quantization collapses in accuracy while LQ-Net
remains robust. Best quantized accuracy per section is \textbf{bolded}. Columns: Acc = Accuracy (\%), Fire = Mean Firing Rate, EMD = EMD from FP32, Dead = Dead Neurons (\%)}
\label{tab:weight_quant_main}
\centering
\small
\setlength{\tabcolsep}{4pt}

\begin{subtable}[t]{0.48\textwidth}
\centering
\caption{CIFAR-100 (FP32: R18: 75.2\%, R8: 72.3\%)}
\begin{tabular}{llrrrr@{\hspace{0.5em}}rrrr}
\toprule
& & \multicolumn{4}{c}{\textbf{ResNet18}} & \multicolumn{4}{c}{\textbf{ResNet8}} \\
\cmidrule(lr){3-6} \cmidrule(lr){7-10}
Method & Bits & Acc & Fire & EMD & Dead & Acc & Fire & EMD & Dead \\
\midrule
\multicolumn{10}{l}{\textit{Narrow Clipping [-1, 1]}} \\
\midrule
Uniform & 2 & 72.8 & .084 & .0166 & 55.3 & 69.3 & .119 & .0161 & 34.5 \\
LQ-Net  & 2 & \textbf{74.1} & .068 & .0118 & 62.0 & 70.9 & .095 & .0066 & 36.8 \\
\addlinespace[0.3em]
Uniform & 4 & 73.9 & .063 & .0113 & 61.5 & 70.8 & .075 & .0056 & 46.2 \\
LQ-Net  & 4 & \textbf{75.3} & .056 & .0074 & 68.0 & \textbf{71.9} & .080 & .0017 & 39.2 \\
\addlinespace[0.3em]
Uniform & 6 & 74.9 & .055 & .0077 & 66.5 & 72.5 & .073 & .0047 & 43.4 \\
LQ-Net  & 6 & \textbf{75.3} & .054 & .0073 & 68.7 & \textbf{72.1} & .077 & .0016 & 38.8 \\
\addlinespace[0.3em]
Uniform & 8 & 74.6 & .048 & .0072 & 71.7 & 72.3 & .078 & .0016 & 40.0 \\
LQ-Net  & 8 & \textbf{75.0} & .049 & .0074 & 72.4 & \textbf{72.2} & .076 & .0014 & 38.4 \\
\midrule
\multicolumn{10}{l}{\textit{Wide Clipping [-10, 10]}} \\
\midrule
Uniform & 2 & 51.6 & .156 & .1206 & 59.8 & 47.5 & .195 & .1518 & 52.1 \\
LQ-Net  & 2 & \textbf{74.4} & .070 & .0121 & 60.3 & \textbf{70.9} & .093 & .0055 & 35.5 \\
\addlinespace[0.3em]
Uniform & 4 & 64.1 & .073 & .0292 & 61.1 & 62.9 & .101 & .0373 & 48.3 \\
LQ-Net  & 4 & \textbf{75.1} & .054 & .0072 & 70.0 & \textbf{71.8} & .082 & .0021 & 37.9 \\
\addlinespace[0.3em]
Uniform & 6 & 73.7 & .071 & .0136 & 59.8 & 69.7 & .103 & .0107 & 38.1 \\
LQ-Net  & 6 & \textbf{75.3} & .049 & .0072 & 71.4 & \textbf{72.1} & .076 & .0016 & 38.1 \\
\addlinespace[0.3em]
Uniform & 8 & 74.4 & .049 & .0083 & 69.9 & 71.5 & .071 & .0060 & 47.0 \\
LQ-Net  & 8 & 75.2 & .053 & .0032 & 71.0 & \textbf{72.6} & .078 & .0012 & 37.4 \\
\bottomrule
\end{tabular}
\end{subtable}
\hfill
\begin{subtable}[t]{0.48\textwidth}
\centering
\caption{CIFAR-10 (FP32: R18: 93.3\%, R8: 91.9\%)}
\begin{tabular}{llrrrr@{\hspace{0.5em}}rrrr}
\toprule
& & \multicolumn{4}{c}{\textbf{ResNet18}} & \multicolumn{4}{c}{\textbf{ResNet8}} \\
\cmidrule(lr){3-6} \cmidrule(lr){7-10}
Method & Bits & Acc & Fire & EMD & Dead & Acc & Fire & EMD & Dead \\
\midrule
\multicolumn{10}{l}{\textit{Narrow Clipping [-1, 1]}} \\
\midrule
Uniform & 2 & 92.1 & .056 & .0103 & 71.4 & 90.0 & .062 & .0113 & 69.3 \\
LQ-Net  & 2 & \textbf{92.9} & .033 & .0028 & 83.9 & \textbf{91.3} & .056 & .0076 & 72.5 \\
\addlinespace[0.3em]
Uniform & 4 & 92.7 & .031 & .0048 & 84.3 & 90.7 & .046 & .0049 & 75.6 \\
LQ-Net  & 4 & \textbf{93.1} & .029 & .0022 & 85.1 & \textbf{91.6} & .048 & .0052 & 74.2 \\
\addlinespace[0.3em]
Uniform & 6 & 93.2 & .029 & .0032 & 85.5 & 91.6 & .046 & .0046 & 73.4 \\
LQ-Net  & 6 & \textbf{93.4} & .027 & .0012 & 86.6 & \textbf{91.9} & .047 & .0047 & 74.4 \\
\addlinespace[0.3em]
Uniform & 8 & 93.4 & .029 & .0025 & 85.5 & 91.9 & .046 & .0045 & 75.4 \\
LQ-Net  & 8 & \textbf{93.1} & .029 & .0018 & 85.2 & \textbf{91.7} & .049 & .0055 & 71.6 \\
\midrule
\multicolumn{10}{l}{\textit{Wide Clipping [-10, 10]}} \\
\midrule
Uniform & 2 & 80.8 & .212 & .1501 & 51.7 & 79.5 & .213 & .1477 & 46.6 \\
LQ-Net  & 2 & \textbf{92.8} & .034 & .0036 & 84.6 & \textbf{91.2} & .055 & .0072 & 71.7 \\
\addlinespace[0.3em]
Uniform & 4 & 91.6 & .059 & .0127 & 70.7 & 89.2 & .088 & .0229 & 57.7 \\
LQ-Net  & 4 & \textbf{93.0} & .029 & .0023 & 85.8 & \textbf{91.9} & .048 & .0050 & 73.0 \\
\addlinespace[0.3em]
Uniform & 6 & 92.6 & .052 & .0106 & 74.1 & 90.6 & .051 & .0069 & 74.6 \\
LQ-Net  & 6 & \textbf{93.4} & .028 & .0013 & 86.1 & \textbf{91.8} & .047 & .0053 & 73.0 \\
\addlinespace[0.3em]
Uniform & 8 & 92.7 & .030 & .0042 & 84.0 & 91.5 & .046 & .0046 & 76.0 \\
LQ-Net  & 8 & \textbf{93.3} & .027 & .0018 & 86.2 & \textbf{91.8} & .047 & .0050 & 74.6 \\
\bottomrule
\end{tabular}
\end{subtable}

\end{table*}

\begin{table}[b!]
\caption{Membrane potential quantization across datasets.
Unsigned ranges ([0,1]) provide more robust dynamics than signed ([-1,1])
across both CIFAR-100 and CIFAR-10, with lower EMD from the FP32 baseline
at equivalent or higher accuracy. Best quantized accuracy per section is bolded. Columns: Acc = Accuracy (\%), Fire = Firing Rate, EMD = EMD from FP32, Dead = Dead Neurons (\%)}
\label{tab:membrane_quant}
\centering
\footnotesize
\setlength{\tabcolsep}{5pt}
\begin{tabular}{llrrrr@{\hspace{1em}}rrrr}
\toprule
& & \multicolumn{4}{c}{\textbf{ResNet18}} & \multicolumn{4}{c}{\textbf{ResNet8}} \\
\cmidrule(lr){3-6} \cmidrule(lr){7-10}
Range & Bits & Acc & Fire & EMD & Dead & Acc & Fire & EMD & Dead \\
\midrule
\multicolumn{10}{l}{\textbf{CIFAR-100} \quad (FP32: R18: 75.2\%, R8: 72.3\%)} \\
\midrule
Signed   & 2 & 67.1 & .041 & .0139 & 82.2 & 66.6 & .063 & .0283 & 67.7 \\
Unsigned & 2 & \textbf{71.9} & .033 & .0104 & 84.9 & \textbf{70.7} & .053 & .0247 & 73.0 \\
\addlinespace
Signed   & 4 & \textbf{74.8} & .073 & .0075 & 67.3 & \textbf{72.4} & .091 & .0132 & 42.2 \\
Unsigned & 4 & 73.3 & .077 & .0097 & 63.3 & 71.4 & .083 & .0143 & 47.7 \\
\addlinespace
Signed   & 6 & \textbf{74.2} & .094 & .0151 & 57.7 & \textbf{71.4} & .117 & .0186 & 33.3 \\
Unsigned & 6 & 73.3 & .077 & .0097 & 62.7 & 71.3& .091 & .0137 & 43.1 \\
\addlinespace
Signed   & 8 & \textbf{73.9} & .115 & .0221 & 54.3 & 72.3 & .120 & .0190 & 31.8 \\
Unsigned & 8 & 73.5 & .089 & .0140 & 62.7 & \textbf{71.6} & .088 & .0141 & 43.4 \\
\midrule
\midrule
\multicolumn{10}{l}{\textbf{CIFAR-10} \quad (FP32: R18: 93.3\%, R8: 91.9\%)} \\
\midrule
Signed   & 2 & 87.2 & .027 & .0090 & 86.6 & 87.3 & .025 & .0123 & 86.8 \\
Unsigned & 2 & \textbf{90.7} & .015 & .0089 & 93.5 & \textbf{89.7} & .024 & .0126 & 88.3 \\
\addlinespace
Signed   & 4 & \textbf{93.1} & .030 & .0050 & 86.0 & 91.5 & .042 & .0068 & 81.1 \\
Unsigned & 4 & 92.8 & .031 & .0033 & 82.6 & \textbf{91.4} & .036 & .0066 & 81.1 \\
\addlinespace
Signed   & 6 & \textbf{93.1} & .045 & .0072 & 77.9 & \textbf{91.6} & .052 & .0064 & 76.4 \\
Unsigned & 6 & 92.4 & .042 & .0053 & 79.0 & 91.3 & .044 & .0064 & 79.1 \\
\addlinespace
Signed   & 8 & \textbf{93.5} & .041 & .0049 & 79.4 & \textbf{92.3} & .051 & .0055 & 74.5 \\
Unsigned & 8 & 92.8 & .042 & .0058 & 77.9 & 90.9 & .041 & .0060 & 79.4 \\
\bottomrule
\end{tabular}
\end{table}

\section{Results}

We evaluate quantization across two primary axes: (1) weight quantization, where network parameters are discretized, and (2) membrane potential quantization, where the recurrent state of LIF neurons is discretized. For each, we assess both accuracy and spiking dynamics across bit-widths (2, 4, 6, 8 bits) and architectures (SEW-ResNet8, SEW-ResNet18) on CIFAR-10 and CIFAR-100.

We find that the LQ-Net approach is effective for weights, and not for membranes. We find that LQ-Net is effective for weight quantization even at very low bitwidths, and that accuracy alone is insufficient to evaluate SNN quantization. We find that methods can preserve performance while degrading firing dynamics in ways that impact deployability and network behavior. We demonstrate this gap quantitatively through Earth Mover's Distance (EMD) and qualitatively through per-neuron firing rate distributions.

\subsection{Weight Quantization Dynamics}
Table~\ref{tab:weight_quant_main} presents weight quantization results across
narrow ([-1, 1]) and wide ([-10, 10]) clipping ranges. The data reveals a
disconnect between accuracy and dynamics preservation.

\subsubsection{Wide Clipping Produces Higher EMD at Similar Accuracy}
\begin{figure}[t]
    \centering
    \includegraphics[width=1\linewidth]{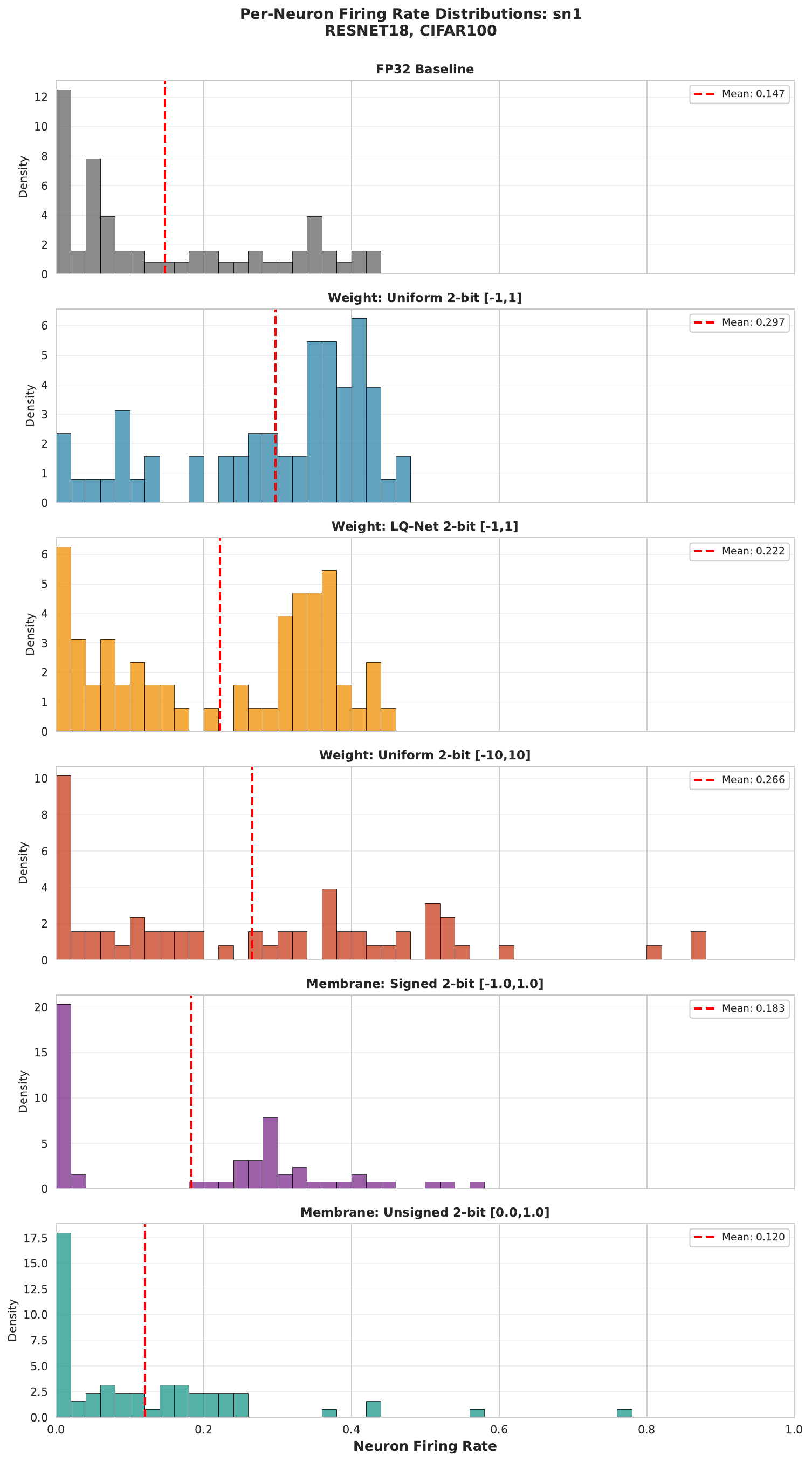}
    \caption{Firing rate distributions across quantization methods at 2-bits
    on SEW-ResNet18 for the CIFAR-100 task for the first spiking layer (sn1).}
    \label{fig:histograms_2bit}
\end{figure}

Figure~\ref{fig:histograms_2bit} shows the per-neuron firing rate distributions
across quantization configurations at 2-bit. Wide-clip uniform quantization
produces a substantially different distribution from the FP32 baseline, with
elevated dead neuron populations and shifted firing rates, while LQ-Net and
narrow-clip uniform quantization remain closer to the full-precision distribution.

This distributional drift is most severe at low bit-widths. At 2-bit with wide
clipping, SEW-ResNet18 on CIFAR-100 drops to 51.6\% accuracy with EMD = 0.121,
reflecting simultaneous accuracy and dynamics degradation. At higher precision
the accuracy gap closes, but EMD remains elevated. At 6-bit, wide-clip uniform quantization
achieves 73.7\% accuracy with EMD = 0.014, while narrow-clip uniform quantization achieves 74.9\%
with EMD = 0.008. A small 1.2\% accuracy difference conceals a 2$\times$
divergence in firing dynamics. On CIFAR-10 ResNet18 at 6-bit, wide-clip uniform quantization
reaches 92.6\% (EMD = 0.011) versus narrow-clip at 93.2\% (EMD = 0.003),
a 3.5$\times$ EMD gap at under 1\% accuracy difference. This is precisely the
regime where accuracy alone can be misleading. The two configurations appear equivalent by standard metrics but differ substantially in firing behavior.

\begin{figure}[h]
    \centering
    \includegraphics[width=1\linewidth]{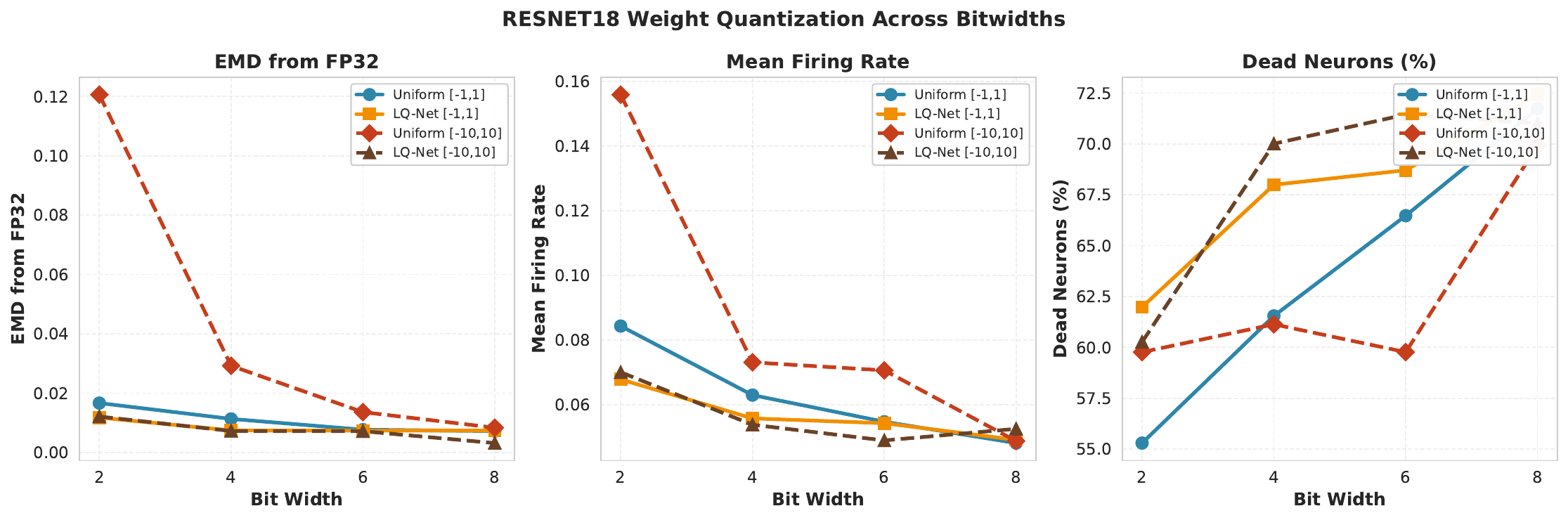}
    \caption{EMD from FP32, mean firing rate, and dead neuron percentage
    across bit-widths for uniform (STE) and learned (LQ-Net) weight
    quantization under narrow [-1,1] and wide [-10,10] clipping on
    SEW-ResNet18 (CIFAR-100).}
    \label{fig:diagnostic_weight}
\end{figure}

Figure~\ref{fig:diagnostic_weight} shows this trend systematically across
bit-widths. Wide-clip uniform quantization exhibits substantially elevated EMD at 2 and 4 bits,
with a correspondingly higher mean firing rate. As precision increases to 6 and
8 bits, accuracy recovers toward the narrow-clip baseline --- but EMD remains
elevated, confirming that firing dynamics lag behind accuracy recovery. LQ-Net
under both clipping ranges maintains low EMD throughout, indicating that
dynamics preservation does not require restricting the clipping range when
quantization levels are learned.

\begin{figure*}[h!]
    \centering
    \includegraphics[width=0.8\linewidth]{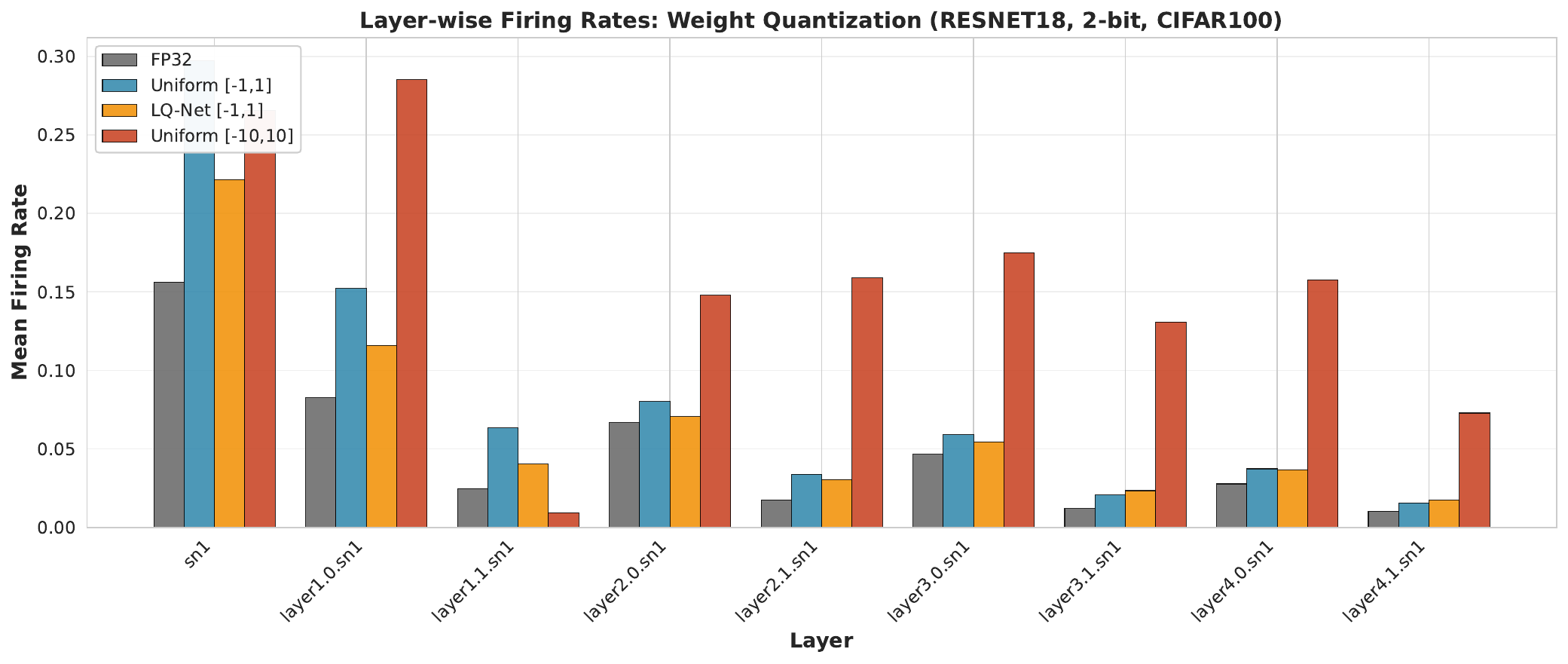}
    \caption{Mean firing rates across layers of trained networks with various weight quantization techniques for 2-bit SEW-ResNet18 trained on the CIFAR-100 task. Layer labels follow SEW-ResNet nomenclature: \texttt{layerX.Y.sn1} 
    denotes the spiking layer in block Y of stage X; the leftmost \texttt{sn1} is the stem spiking layer.}
    \label{fig:weight_layer_barchart}
\end{figure*}

\begin{figure}[h!]
    \centering
    \includegraphics[width=1\linewidth]{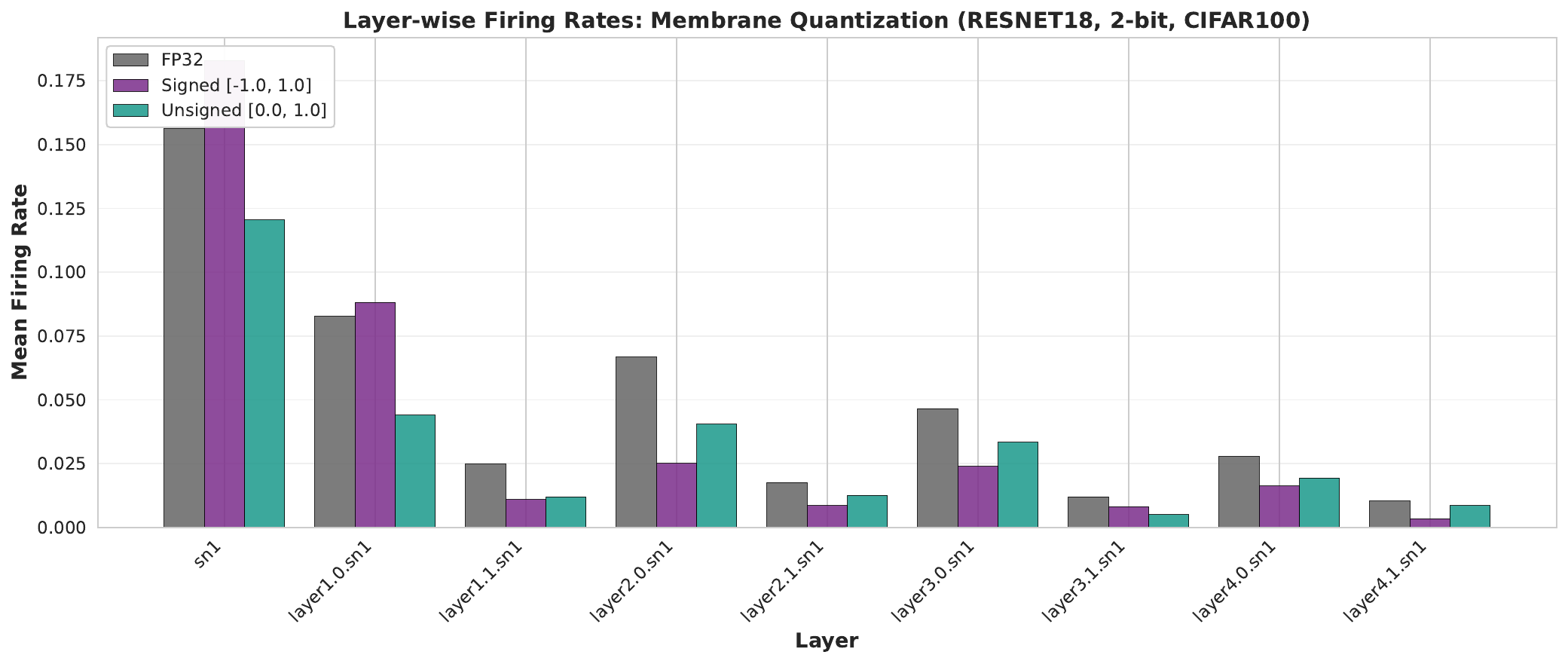}
    \caption{Mean firing rates across layers of trained networks with membrane quantization techniques for 2-bit SEW-ResNet18 trained on the CIFAR-100 task. Layer labels follow SEW-ResNet nomenclature: \texttt{layerX.Y.sn1} denotes the spiking layer in block Y of stage X; the leftmost \texttt{sn1} is the stem spiking layer.}
    \label{fig:membrane_layer_barchart}
\end{figure}

The pattern holds across architectures and datasets. On CIFAR-10 ResNet8 at 2-bit, wide-clip uniform quantization achieves 79.5\% accuracy with EMD = 0.148, versus 90.0\% and EMD = 0.011 for narrow clipping, a 13$\times$ EMD divergence alongside a substantial accuracy gap. At 4-bit the accuracy gap narrows to 1.5\%, but EMD remains 4$\times$ higher under wide clipping (0.023 versus 0.005). This consistent pattern across ResNet8 and ResNet18 on both datasets confirms that the dynamics-accuracy disconnect is a systematic property of wide clipping under
uniform quantization, not an artifact of a single configuration.

\subsubsection{Learned Quantization Preserves Both Accuracy and Dynamics}
LQ-Net outperforms uniform quantization on both metrics. At 2-bit
with wide clipping (CIFAR-100, ResNet18), where uniform quantization collapses to
51.6\%, LQ-Net achieves 74.4\% accuracy with EMD = 0.012. It maintains dynamics
close to the FP32 baseline while uniform quantization produces EMD = 0.121, a
10$\times$ divergence. The firing rate distribution (Figure~\ref{fig:histograms_2bit})
confirms this: LQ-Net remains near the FP32 baseline across the firing rate range,
while wide-clip STE shows substantial drift.

LQ-Net's advantage is most pronounced precisely where uniform quantization fails: wide clipping at low bit-widths. Table~\ref{tab:weight_quant_main} shows this trend across bit-widths. For narrow clipping, both methods achieve similar accuracy (within 1--2\%) and comparable EMD. With wide clipping, uniform quantization degrades substantially in both accuracy and dynamics at 2 and 4 bits, while LQ-Net remains robust across all configurations. As expected from its design, LQ-Net is largely invariant to clipping range since the quantization levels are learned rather than fixed.

\begin{figure}[h]
    \centering
    \includegraphics[width=1\linewidth]{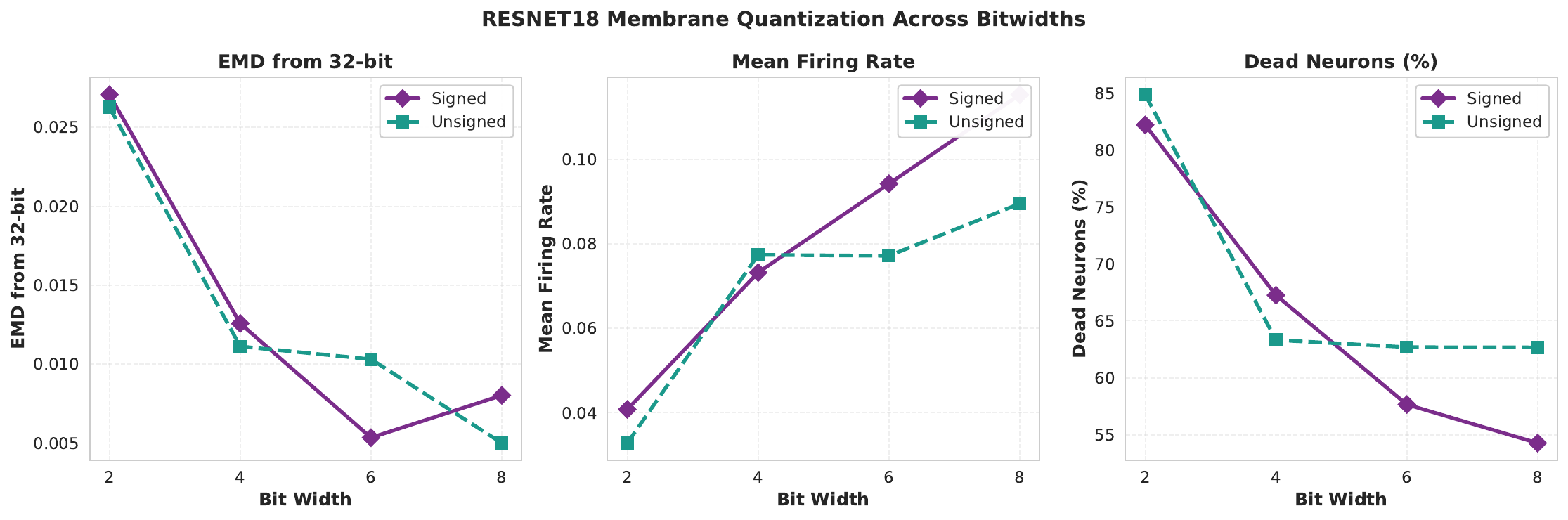}
    \caption{EMD, dead neuron percentage, and mean firing rate across 
    bit-widths for signed and unsigned membrane quantization on 
    SEW-ResNet18 (CIFAR-100). All three metrics converge toward 
    full-precision levels as precision increases, supporting EMD 
    as a diagnostic for firing behavior recovery.}
    \label{fig:diagnostic_membrane}
\end{figure}

\subsubsection{Layer-wise Analysis Reveals Systematic Degradation}
Figure~\ref{fig:weight_layer_barchart} shows firing rates across layers for ResNet18 on CIFAR-100. Wide-clip uniform quantization exhibits systematically elevated firing rates across layers compared to both narrow-clip STE and LQ-Net, which remain close to the FP32 baseline. Since firing activity directly can determine the number of spike-driven accumulations, elevated firing rates may translate to increased computational load at inference time, independent of accuracy. LQ-Net maintains firing rates close to the FP32 baseline network-wide, suggesting that dynamics-preserving quantization retains not just behavioral fidelity but also the sparsity structure that makes SNNs efficient in the first place.

\subsection{Membrane Potential Quantization}
Membrane quantization introduces challenges distinct from weight quantization. Since membrane potentials are recurrent states updated at every timestep, quantization errors compound temporally through the LIF dynamics. Table~\ref{tab:membrane_quant} presents results for signed ([-1.0, 1.0]) versus unsigned ([0.0, 1.0]) ranges across CIFAR-10 and CIFAR-100. We find that naively applying LQ-Net to membrane
quantization results in training collapse, leaving uniform quantization as the only viable approach in this setting.

\subsubsection{Unsigned Ranges Provide More Robust Dynamics}
At 2-bit, unsigned quantization provides both higher accuracy and lower EMD than signed ranges. For SEW-ResNet18 on CIFAR-100, unsigned achieves 71.9\% accuracy with EMD = 0.010, while signed yields 67.1\% with EMD = 0.014. This pattern holds on CIFAR-10: unsigned reaches 90.7\% versus signed's 87.2\%, with comparable EMD. The unsigned advantage at 2-bit is consistent across both ResNet8 and ResNet18, suggesting it reflects a fundamental property of the quantization range rather than architecture-specific behavior.

At higher bit-widths the picture is more nuanced. At 4-bit and above, signed and unsigned achieve similar accuracy on both datasets, with EMD differences that are small in absolute terms. Signed ranges occasionally achieve marginally higher accuracy at 4--8 bits (e.g. 74.8\% versus 73.3\% for ResNet18 CIFAR-100 at 4-bit), while unsigned tends to produce slightly lower EMD at low bit widths. Figure~\ref{fig:diagnostic_membrane} shows these trends across bit-widths: EMD decreases for both ranges as precision increases, with unsigned consistently tracking closer to the FP32 baseline in firing dynamics even when accuracy is comparable. 

The firing rate distributions in Figure~\ref{fig:histograms_2bit} illustrate the 2-bit case directly. Signed membrane quantization produces compressed firing rates in later layers and elevated dead neuron percentages, while unsigned maintains a distribution closer to FP32 at the same bit-width.

\subsubsection{Membrane Quantization Requires Higher Precision}
Unlike weight quantization, where LQ-Net achieves competitive accuracy even at 2-bit, membrane quantization requires $\geq$4 bits for reliable accuracy and dynamics preservation under both signed and unsigned ranges. At 4-bit, both approaches achieve accuracy within 1\% of FP32 on CIFAR-100 and CIFAR-10, with EMD below 0.015. At 2-bit, accuracy drops 4--8\% and EMD exceeds 0.010, indicating fundamental dynamics disruption.

The precision sensitivity may stem from timesteps and non stationary input for membrane quantization. A weight quantization error is fixed for an entire forward pass, whereas a membrane quantization error propagates through the recurrent LIF update at every timestep, potentially amplifying over the $T=4$ timesteps we use. This compounding effect makes membrane quantization inherently more sensitive to precision than weight
quantization, and may explain why the LQ-Net training instability we observe is specific to membranes.

\subsubsection{Layer-wise Analysis of Membrane Potentials}
Figure~\ref{fig:membrane_layer_barchart} shows per-layer firing rates for 2-bit
membrane quantization on ResNet18 CIFAR-100. Both signed and unsigned quantization
produce lower mean firing rates than FP32 across most layers, reflecting the
dynamics compression introduced at this precision. Unsigned quantization maintains
firing rates closer to FP32 in most layers. While both achieve reasonable accuracy, unsigned is able to outperform signed at 2 bit (see Table~\ref{tab:membrane_quant}), while also firing less, especially at the firing layer. Similar to the weight quantization setting, the act of quantization induces structural changes in the firing behavior of network across layers.

\subsection{EMD as a Diagnostic Metric}
Figures~\ref{fig:diagnostic_weight} and~\ref{fig:diagnostic_membrane} demonstrate 
that EMD tracks dynamics degradation systematically across bit-widths for both 
weight and membrane quantization. For weight quantization (Figure~\ref{fig:diagnostic_weight}), 
wide-clip uniform quantization exhibits elevated EMD at 2 and 4 bits, with a corresponding elevation in mean firing rate. As precision increases to 6 and 8 bits, accuracy recovers toward the narrow-clip baseline. As expected, EMD of uniform quantization remains elevated relative to LQ-Net. 

For membrane quantization (Figure~\ref{fig:diagnostic_membrane}), EMD decreases 
with increasing precision for both signed and unsigned ranges, alongside 
corresponding recovery in dead neuron percentage and mean firing rate toward 
full-precision levels. This coherence across all three metrics suggests EMD is 
capturing genuine distributional recovery rather than incidental variation. No matter the distributional shift between the behavior of any two models, EMD represents a one number metric to capture their behavioral similarity.

\section{Conclusion}
Quantization evaluation in SNNs has focused almost exclusively on accuracy,
but accuracy alone does not capture whether a quantized network behaves like
its full-precision counterpart. Our experiments show that quantization method,
clipping range, and bit-width can produce substantially different firing
distributions at equivalent accuracy. These differences are invisible to accuracy alone, but are relevant to how a network behaves in deployment.

Earth Mover's Distance provides a principled way to surface these differences.
Across architectures and datasets, EMD tracks distributional shifts in firing
behavior that mean firing rate and dead neuron percentage alone can miss.
At 6-bit with wide clipping, for instance, uniform quantization achieves
accuracy within 1--2\% of narrow clipping but produces up to 3.5$\times$
higher EMD, revealing distributional drift that scalar summaries obscure
(Figure~\ref{fig:histograms_2bit}, Table~\ref{tab:weight_quant_main}).
Aas precision increases, accuracy recovers faster than EMD
under wide-clip uniform quantization, indicating that dynamics lag behind accuracy recovery and that accuracy alone underestimates residual distributional drift (Figure~\ref{fig:diagnostic_weight}). Learned quantization largely resolves this: LQ-Net maintains firing distributions close to the full-precision baseline across configurations where uniform quantization degrades, demonstrating
that behavior-preserving quantization is achievable without sacrificing task performance. Membrane quantization presents a distinct challenge, errors
compound across timesteps through the recurrent LIF dynamics, requiring $\geq$4 bits for reliable dynamics preservation and favoring unsigned ranges at low bit-widths where both accuracy and EMD are superior (Table~\ref{tab:membrane_quant}, Figure~\ref{fig:diagnostic_membrane}).

Taken together, these results suggest that behavior preservation should be treated as an important evaluation criterion alongside accuracy for quantized SNNs. EMD offers a practical diagnostic for this purpose, and we encourage its adoption when assessing whether quantization choices preserve network behavior. 

\bibliographystyle{ACM-Reference-Format}
\bibliography{references}

\end{document}